\documentclass[a4paper,conference]{IEEEtran}
\ifCLASSINFOpdf
\else
\fi
\usepackage{amsmath,amssymb,amsthm, bm}
\usepackage[dvipdfmx]{graphicx}
\usepackage{enumerate}
\usepackage{my-style}
\allowdisplaybreaks[4]
\graphicspath{{fig/}}

\begin{document}
%
\title{Multi-layered Discriminative Restricted Boltzmann Machine with Untrained Probabilistic Layer}

\author{\IEEEauthorblockN{Yuri Kanno}
\IEEEauthorblockA{Graduate School of Science and Engineering,\\Yamagata University, Japan}
\and
\IEEEauthorblockN{Muneki Yasuda}
\IEEEauthorblockA{Graduate School of Science and Engineering,\\Yamagata University, Japan}
}

\maketitle

\begin{abstract}
An extreme learning machine (ELM) is a three-layered feed-forward neural network having untrained parameters,  
which are randomly determined before training. 
Inspired by the idea of ELM, a probabilistic untrained layer called a probabilistic-ELM (PELM) layer is proposed, 
and it is combined with a discriminative restricted Boltzmann machine (DRBM), which is a probabilistic three-layered neural network for solving classification problems. 
The proposed model is obtained by stacking DRBM on the PELM layer.
The resultant model (i.e., multi-layered DRBM (MDRBM)) forms a probabilistic four-layered neural network. 
In MDRBM, the parameters in the PELM layer can be determined using Gaussian-Bernoulli restricted Boltzmann machine. 
Owing to the PELM layer, MDRBM obtains a strong immunity against noise in inputs, 
which is one of the most important advantages of MDRBM. 
Numerical experiments using some benchmark datasets, 
MNIST, Fashion-MNIST, Urban Land Cover, and CIFAR-10, 
demonstrate that MDRBM is superior to other existing models, particularly, 
in terms of the noise-robustness property (or, in other words, the generalization property).
\end{abstract}


%
\IEEEpeerreviewmaketitle

\section{Introduction}

A discriminative restricted Boltzmann machine (DRBM)~\cite{DRBM2012} is a probabilistic three-layered neural network for solving classification problems. 
DRBM is constructed based on a restricted Boltzmann machine (RBM)~\cite{RBM1986,CD2002}. 
The inference and learning in general Boltzmann machines are computationally difficult 
because they involve exponential computational costs. 
On the other hand, those in DRBM can be exactly performed owing to its special structure. 
The main aim of this study is to propose a probabilistic four-layered neural network by multi-layering DRBM. 
A possible extension of DRBM is a deep Boltzmann machine (DBM)~\cite{DBM2009} or its variants such as Gaussian-Bernoulli deep Boltzmann machine~\cite{DeepGBRBM2013}. 
However, unlike DRBM, the inference and learning in DBM cannot be performed exactly, 
and approximate algorithms for them often need rather expensive sampling procedures based on the Markov chain Monte Carlo method. 
Furthermore, the vanishing-gradient problem~\cite{DL2016} cannot be ignored in multi-layering based on DBM without an appropriate pretraining procedure. 

An extreme learning machine (ELM) is basically a three-layered feed-forward neural network~\cite{ELM2006}. 
Unlike the standard feed-forward neural network, only the parameters between the output and hidden layers are trained during training in ELM. 
The remaining untrained parameters, between the hidden and input layers, are randomly determined before training. 
Because the number of tuned parameters of ELM is less than that of the standard feed-forward neural network, 
the representation power of ELM is less than that of the corresponding feed-forward neural network in general. 
However, ELM tends to be more robust, or more generalized in other words, owing to this property.  

Based on the idea of ELM, DRBM is extended as follows. 
A probabilistic untrained layer based on the idea of ELM, which is called a probabilistic-ELM (PELM) layer in this study, 
is introduced and it is combined with DRBM (i.e., DRBM is stacked on the PELM layer). 
The resultant model forms a probabilistic four-layered neural network. 
Although the inference and learning in the proposed model needs a sampling approximation, it is not expensive. 
Moreover, the vanishing-gradient problem will not be severe, because the parameters in the additive layer are untrained. 
The parameters in the PELM layer are determined using Gaussian-Bernoulli restricted Boltzmann machine (GBRBM)~\cite{RBM-Sci2006,GBRBM2011}. 
The idea of this determination is very similar to that in reference \cite{ELM-GBRBM2018}.
By the PELM layer constructed from GBRBM, the proposed probabilistic four-layered neural network obtains a strong immunity against noise in inputs. 
In the proposed model, the additive probabilistic layer sends the fluctuating signals to the stacked DRBM. 
This scheme provides important effects to both learning and inference states: 
it functions as a data augmentation in the learning stage; and while, it can suppress the increase in the variances of the signals caused by the input noise.
The robustness for input noise is one of the most important advantages of the proposed model. 
Numerical experiments using some benchmark datasets demonstrate that the proposed model is superior to DRBM and the other existing models~\cite{ELM-GBRBM2018,CDRBM2019}, 
particularly, in terms of the noise-robustness property (in other words, the generalization property). 

The remainder of this paper is organized as follows. 
In section \ref{sec:DRBM}, a brief introduction to DRBM in presented. 
The proposed model is introduced in section \ref{sec:Proposed}. 
In section \ref{sec:Using_GBRBM}, the setting of the untrained parameters in the proposed model is discussed, 
and a method based on pretraining-like unsupervised learning using GBRBM is proposed.
Section \ref{sec:experiment} presents the validity of the proposed model through numerical experiments using some benchmark datasets: 
MNIST, Fashion-MNIST (F-MNIST), Urban Land Cover (ULC)~\cite{ULC2013}, and CIFAR-10.
Finally, a summary and discussions are given in section \ref{sec:conclusion}.

\section{Discriminative Restricted Boltzmann Machine}
\label{sec:DRBM}

Consider a classification problem in which an $n$-dimensional input vector $\bm{x} := (x_1, x_2,\ldots, x_n)^{\mrm{T}} \in \mathbb{R}^n$
is classified into $K$ different classes $C_1, C_2, \ldots, C_K$. 
It is convenient to use the 1-of-$K$ vector (or the one-hot vector) to identify each class~\cite{Bishop2006}, 
where each class corresponds to the $K$-dimensional vector $\bm{t} := (t_1, t_2,\ldots, t_K)^{\mrm{T}}$ having the elements 
$t_k \in \{0,1\}$ and $\sum_{k = 1}^K t_k = 1$, i.e., a vector in which only one element is one and the remaining elements are zero. 
When $t_k = 1$, $\bm{t}$ indicates class $C_k$. 
For the sake of simplicity, the 1-of-$K$ vector, the $k$th element of which is one, is denoted by $\bm{1}_k$,  
so that $\bm{t} \in T_K := \{\bm{1}_k \mid k = 1,2,\ldots,K\}$. 
Thus, $\bm{t} = \bm{1}_k$ corresponds to class $C_k$. 

\begin{figure}[tb]
\centering
\includegraphics[height=3cm]{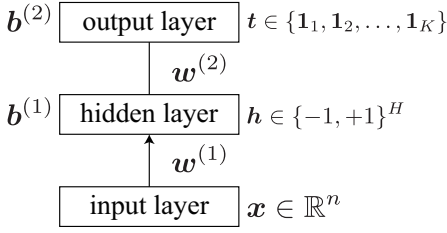}
\caption{Illustration of DRBM as a probabilistic three-layered neural network. 
$\bm{b}^{(1)}$ and $\bm{b}^{(2)}$ are the biases in the hidden layer and output layer, respectively. 
$\bm{w}^{(1)}$ and $\bm{w}^{(2)}$ are the connection weights between the input and hidden layers and between the hidden and output layers, respectively.
The undirected connections, $\bm{w}^{(2)}$, represent the interactive connections.}
\label{fig:DRBM}
\end{figure}

DRBM was proposed to solve the classification problem~\cite{DRBM2012}. 
It is a probabilistic three-layered neural network in which the input layer consist of $n$ input variables $\bm{x}$, 
hidden layer consist of $H$ hidden variables $\bm{h} :=(h_1, h_2,\ldots, h_H)^{\mrm{T}}$, 
and the output layer consists of $K$ output variables $\bm{t}$ (see figure \ref{fig:DRBM}). 
Each hidden variable takes $h_j \in \{-1,+1\}$. 
In DRBM, the joint distribution over $\bm{t} \in T_K$ and $\bm{h} \in \{-1,+1\}^H$ is expressed by 
\begin{align}
&P(\bm{t},\bm{h} \mid \bm{x}, \theta):= \frac{1}{Z(\bm{x},\theta)}
\exp\big(-E(\bm{t},\bm{h}; \bm{x}, \theta)\big),
\label{eqn:DRBM}
\end{align}
where $E(\bm{t},\bm{h}; \bm{x}, \theta)$ is the energy function defined by
\begin{align*}
E(\bm{t},\bm{h}; \bm{x}, \theta)&:=-\sum_{k=1}^K b_{k}^{(2)} t_k - \sum_{j = 1}^H b_{j}^{(1)} h_j
-\sum_{k=1}^K\sum_{j=1}^H w_{k,j}^{(2)}t_kh_j\nn
\aldef
- \sum_{j=1}^H\sum_{i=1}^n w_{j,i}^{(1)} h_jx_i, 
\end{align*}
and $Z(\bm{x},\theta)$ is the partition function.
Here, the set of learning parameters in DRBM, i.e., $\bm{b}^{(1)}$, $\bm{b}^{(2)}$, $\bm{w}^{(1)}$, and $\bm{w}^{(2)}$, is collectively denoted by the set $\theta$.

In the inference of DRBM, the conditional distribution of DRBM is used, which is obtained by marginalizing $\bm{h}$ out from equation (\ref{eqn:DRBM}):   
\begin{align}
P(\bm{t} \mid \bm{x}, \theta) = \sum_{\bm{h}}P(\bm{t},\bm{h} \mid \bm{x}, \theta)
\label{eqn:classProb_DRBM}
\end{align}
where $\sum_{\bm{h}}$ is the sum over all the possible realizations of $\bm{h}$, i.e., 
$\sum_{\bm{h}} = \prod_{j=1}^H \sum_{h_j \in \{-1,+1\}}$. 
$P(\bm{t} \mid \bm{x}, \theta)$ denotes the class probability; for example,
$P(\bm{1}_k \mid \bm{x}, \theta)$ is the class probability of $C_k$. 
The input $\bm{x}$ is categorized into the class with the maximum probability.
The computational cost of evaluating all the class probabilities for a certain $\bm{x}$ is $O(H(n + K))$.  

The learning parameters of DRBM, $\theta$, are optimized by using the maximum-likelihood estimation (MLE). 
That is, for a training dataset consisting of $N$ training data points $\mcal{D}:=\{(\mbf{x}^{(\mu)}, \mbf{t}^{(\mu)} )\mid \mu = 1,2,\ldots, N\}$, 
the log likelihood of equation (\ref{eqn:classProb_DRBM}),
$\ell(\theta):= N^{-1}\sum_{\mu =1}^N \ln P(\mbf{t}^{(\mu)} \mid \mbf{x}^{(\mu)}, \theta)$,  
is maximized with respect to $\theta$. 
The exact inference and learning cannot be executed in RBM, because they involve an exponential computational cost. 
In contrast, they can be executed in DRBM~\cite{DRBM2012}. 
In fact, in DRBM, the inference cost for a certain input is $O(H(n + K))$, as mentioned above,  
and the learning cost (i.e., the cost of evaluating all the gradients of $\ell(\theta)$) is $O(NH(n + K))$. 

\section{Proposed Model}
\label{sec:Proposed}

ELM is a three-layered feed-forward neural network~\cite{ELM2006}. 
In a standard feed-forward neural network, all the parameters in the system (the biases and connection weights) are trained by using a given dataset. 
In contrast, in ELM, only the connection weights between the hidden and output layers and the biases in the output layer are trained during training, 
and the remaining untrained parameters (i.e., the connection weights between the input and hidden layers and the biases in the hidden layer) 
are randomly determined before training. 
With an appropriate setting for the untrained parameters, 
the results obtained using ELM can be comparable to those obtained using standard feed-forward neural networks~\cite{ELM2006}. 
The number of tuned parameters of ELM is less than the corresponding feed-forward neural network.
Therefore, the representation power of ELM is less than that of the corresponding feed-forward neural network. 
However, ELM tends to be more robust, or in other words, more generalized.  
Hence, ELM can be effective in disadvantageous situations, such as situations in which the training dataset is small or input data are noisy. 

In the following, we extend the idea of ELM to a probabilistic model and then consider a combination of it with DRBM. 

\subsection{Untrained probabilistic layer: probabilistic-ELM layer}
\label{sec:PELM}

In ELM, the feed-forward signal to the $j$th hidden unit from the input layer is expressed as
$u_j(\bm{x}) := b_j^{(0)} + \sum_{i=1}^n w_{j,i}^{(0)} x_i$,
where $\bm{b}^{(0)}$ denotes the biases of the hidden layer and $\bm{w}^{(0)}$ denotes the (directed) connection weights between the input layer and hidden layer. 
The output signal of the $j$th hidden unit is $z_j = \mrm{act}(u_j(\bm{x}))$, where $\mrm{act}(u)$ is a certain activation function. 

Based on the above feed-forward propagation, a probabilistic layer expressed by the Bernoulli distribution 
over $\bm{z}:=(z_1, z_2,\ldots, z_{\mcal{H}})^{\mrm{T}} \in \{-1,+1\}^{|\mcal{H}|}$ is defined as
\begin{align}
B(\bm{z} \mid \bm{x}, \theta_0):= \prod_{j=1}^{\mcal{H}}\frac{\exp ( u_j(\bm{x}) z_j)}{2 \cosh u_j(\bm{x})},
\label{eqn:PELM_layer}
\end{align}
where $\theta_0$ is the set of the parameters in $\bm{b}^{(0)}$ and $\bm{w}^{(0)}$.
The units $\bm{z}$ in this probabilistic layer stochastically take a value of $-1$ or $+1$ according to the Bernoulli distribution in equation (\ref{eqn:PELM_layer}). 
This probabilistic layer is referred to as the PELM layer.  

\subsection{Multi-layered discriminative restricted Boltzmann machine}
\label{sec:MDRBM}

\begin{figure}[tb]
\centering
\includegraphics[height=4cm]{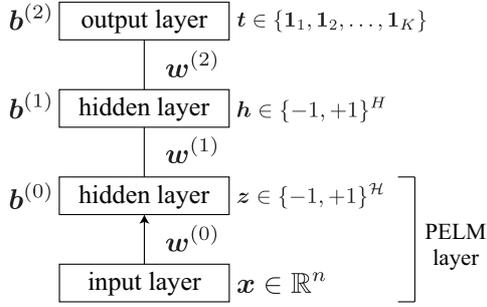}
\caption{Illustration of MDRBM as a four-layered probabilistic neural network. $\bm{t}$ and $\bm{h}$, and $\bm{z}$ are all random variables. 
$\bm{b}^{(0)}$ and $\bm{w}^{(0)}$ are the untrained parameters that are determined before training.}
\label{fig:MDRBM}
\end{figure}

The PELM layer introduced in the previous section is combined with DRBM as
\begin{align}
P^{\dagger}(\bm{t},\bm{h}, \bm{z} \mid \bm{x}, \theta, \theta_0) 
:=P(\bm{t},\bm{h} \mid \bm{z}, \theta) B(\bm{z} \mid \bm{x}, \theta_0).
\label{eqn:JointDist_MDRBM}
\end{align}
Here, the first factor on the right-hand side of equation (\ref{eqn:JointDist_MDRBM}) is DRBM, 
and the second factor is the PELM layer defined in equation (\ref{eqn:PELM_layer}). 
The joint distribution $P^{\dagger}(\bm{t},\bm{h}, \bm{z} \mid \bm{x}, \theta, \theta_0)$ expresses the four-layered probabilistic neural network, 
referred to as the multi-layered DRBM (MDRBM), as shown in figure \ref{fig:MDRBM}.
In MDRBM, $\theta_0$ is regarded as the untrained parameter set that is determined before training, in accordance with the concept of ELM. 

The class probability in MDRBM is expressed by
\begin{align}
P^{\dagger}(\bm{t} \mid \bm{x}, \theta, \theta_0)
&= \sum_{\bm{z}}P(\bm{t} \mid \bm{z}, \theta)B(\bm{z} \mid \bm{x}, \theta_0),
\label{eqn:classProb_MDRBM}
\end{align}
where $\sum_{\bm{z}}$ is the sum over all the possible realizations of $\bm{z}$. 
$P(\bm{t} \mid \bm{z}, \theta)$ in equation (\ref{eqn:classProb_MDRBM}) is the class probability of DRBM described in equation (\ref{eqn:classProb_DRBM}). 
Because the sum over $\bm{z}$ cannot be taken owing to its computational-cost requirement, 
it is evaluated by a sample approximation (or Monte Carlo integration):
\begin{align}
P^{\dagger}(\bm{t} \mid \bm{x}, \theta, \theta_0)
&\approx \frac{1}{S}\sum_{\nu=1}^SP(\bm{t} \mid \bm{z}^{(\nu)}, \theta),
\label{eqn:classProb_MDRBM_MCI}
\end{align}
where $z^{(1)}, z^{(2)},\ldots,z^{(S)}$ are $S$ sample points drawn from the distribution $B(\bm{z} \mid \bm{x}, \theta_0)$. 
It is noteworthy that sampling from $B(\bm{z} \mid \bm{x}, \theta_0)$ is very easy because all $z_j$s are statistically independent of each other. 
In MDRBM, the computational cost of the inference for a certain input is $O(SH(\mcal{H} + K) + n\mcal{H})$, 
where, $SH(\mcal{H} + K)$ originates from the computation of equation (\ref{eqn:classProb_MDRBM_MCI}) and 
$n\mcal{H}$ originates from the computation of equation (\ref{eqn:PELM_layer}).

The learning of MDRBM is achieved by MLE, i.e., by maximizing the log likelihood of equation (\ref{eqn:classProb_MDRBM}),  
\begin{align}
\ell^{\dagger}(\theta;\theta_0):=\frac{1}{N}\sum_{\mu=1}^N \ln P^{\dagger}(\mbf{t}^{(\mu)} \mid \mbf{x}^{(\mu)}, \theta, \theta_0),
\label{eqn:likelihood_MDRBM}
\end{align}  
with respect to $\theta$. 
The log likelihood is maximized using a gradient ascent method. 
The gradient of $\ell^{\dagger}(\theta;\theta_0)$ with respect to a learning parameter $\alpha \in \theta$ is obtained as
\begin{align}
&\frac{\partial \ell^{\dagger}(\theta;\theta_0)}{\partial \alpha} \nn
&=\frac{1}{N}\sum_{\mu=1}^N  \sum_{\bm{z}}\frac{\partial \ln P(\mbf{t}^{(\mu)} \mid \bm{z}, \theta)}{\partial \alpha}
P^{\dagger}(\bm{z}\mid \mbf{t}^{(\mu)}, \mbf{x}^{(\mu)}, \theta, \theta_0),
\label{eqn:grad_likelihood_MDRBM}
\end{align}
where
\begin{align*}
P^{\dagger}(\bm{z}\mid \bm{t}, \bm{x}, \theta, \theta_0)=
\frac{P(\bm{t} \mid \bm{z}, \theta)B(\bm{z} \mid \bm{x}, \theta_0)}
{P^{\dagger}(\bm{t} \mid \bm{x}, \theta, \theta_0)}
\end{align*}
is the conditional distribution of $\bm{z}$, given $\bm{x}$ and $\bm{t}$. 
Because the gradient in equation (\ref{eqn:grad_likelihood_MDRBM}) involves an intractable sum over $\bm{z}$, 
it is evaluated using the sample approximation in a manner similar to equation (\ref{eqn:classProb_MDRBM_MCI}): 
\begin{align}
&\frac{\partial \ell^{\dagger}(\theta;\theta_0)}{\partial \alpha} \nn
&\approx\frac{1}{N S}\sum_{\mu=1}^N  \sum_{\nu = 1}^S \frac{\partial \ln P(\mbf{t}^{(\mu)} \mid \bm{z}^{(\mu, \nu)}, \theta)}{\partial \alpha}
\frac{P(\mbf{t}^{(\mu)} \mid \bm{z}^{(\mu,\nu)}, \theta)}
{P^{\dagger}(\mbf{t}^{(\mu)} \mid \mbf{x}^{(\mu)}, \theta, \theta_0)},
\label{eqn:grad_likelihood_MDRBM_MCI}
\end{align}
where $z^{(\mu, 1)}, z^{(\mu, 2)},\ldots,z^{(\mu, S)}$ are $S$ sample points drawn from $B(\bm{z} \mid \mbf{x}^{(\mu)}, \theta_0)$. 
Thus, the learning cost of MDRBM is $O(N\{SH(\mcal{H} + K) + n\mcal{H}\})$. 

In MDRBM, the stacked DRBM is received the fluctuating signals from the PELM layer. 
This scheme can be viewed as a data-augmentation scheme in the training; 
and therefore, this can help to increase the strength of noise robustness. 
Furthermore, the fluctuating signals also can help to increase the strength of noise robustness in the inference;  
the detailed discussion of it is presented in section \ref{sec:discussion}

\section{Determination of Untrained Parameters} 
\label{sec:Using_GBRBM}

In MDRBM, it is assumed that the parameters in $\theta_0$ are untrained ones, 
the values of which are determined before training. 
The values of the parameters in $\theta_0$ are very important 
because they directly affect the feature mapping from the input layer to the first hidden layer. 
In ELM, such untrained parameters are determined randomly. 
This type of strategy can be also employed in MDRBM. 

In the following, an alternative strategy based on GBRBM~\cite{RBM-Sci2006,GBRBM2011} is considered. 
GBRBM is defined in the following form: 
\begin{align}
G(\bm{z}, \bm{x} \mid \theta_0, \theta_1)
&:=\frac{1}{Z_G(\theta_0, \theta_1)}
\exp\big( - E_G(\bm{z}, \bm{x}; \theta_0, \theta_1)\big), 
\label{eqn:GBRBM} \\
E_G(\bm{z}, \bm{x}; \theta_0, \theta_1)&:=
\frac{1}{2}\sum_{i=1}^n \frac{x_i^2}{\sigma_i^2} - \sum_{i=1}^n c_i x_i
-\sum_{j = 1}^{\mcal{H}}b_j^{(0)}z_j \nn
\aldef
-\sum_{j=1}^{\mcal{H}}\sum_{i=1}^n w_{j,i}^{(0)} z_jx_i,
\end{align}  
where $\theta_1$ is the set of the parameters, $\bm{\sigma}$ and $\bm{c}$,   
and $Z_G(\theta_0, \theta_1)$ is the partition function.
In GBRBM, $\bm{x}$ are regarded as the visible variables and $\bm{z}$ are regarded as the hidden variables.  
Note that the definition in equation (\ref{eqn:GBRBM}) is slightly modified from the original ones~\cite{RBM-Sci2006,GBRBM2011} 
but is essentially the same. 
The conditional  distribution of GBRBM is equivalent to the Bernoulli distribution in equation (\ref{eqn:PELM_layer}): 
$G(\bm{z}  \mid \bm{x}, \theta_0, \theta_1) = B(\bm{z} \mid \bm{x}, \theta_0)$.
The learning of GBRBM is achieved by maximizing the marginal log likelihood, 
$\ell_G(\theta_0, \theta_1):= N^{-1}\sum_{\mu = 1}^N \sum_{\bm{z}}G(\bm{z}, \mbf{x}^{(\mu)} \mid \theta_0, \theta_1)$,
with respect to $\theta_0$ and $\theta_1$. 
This GBRBM-learning is unsupervised because it uses only input data in the given dataset. 
The maximization of $\ell_G(\theta_0, \theta_1)$ can be performed using the contrastive divergence method~\cite{CD2002}. 
The untrained parameters in $\theta_0$, trained using the GBRBM-learning method, is used in MDRBM. 

\subsection{Related models}

In the original DRBM, the hidden variables are binary variables.
Recently, an extended DRBM, called continuous-DRBM (CDRBM), has been proposed~\cite{CDRBM2019} 
for the purpose of the improvement of the generalization property. 
In CDRBM, the hidden variables are treated as $[-1,+1]$-continuous variables.
CDRBM has been numerically proven to be superior to the original DRBM in terms of the generalization property 
in a bad condition such as the case of the size of training dataset is strongly limited or the case of the training and test datasets being dissimilar to each other. 
The continuous hidden variables can help to avoid data over-fitting in such a case. 

The idea of the proposed model, MDRBM with the GBRBM-learning method, is the same as that of RBM-ELM~\cite{ELM-GBRBM2018}. 
In RBM-ELM, the untrained parameters of ELM are determined via the GBRBM-learning method. 
It was shown that RBM-ELM can achieve a better performance than the standard ELM and its variant models. 

In the experiments in the following section, the proposed model is compared with these related models.

\section{Numerical Experiments}
\label{sec:experiment}

In this section, results of numerical experiments are presented. 
All the input data used in the following experiments were standardized in the preprocessing, i.e., the Z-score normalization. 
In the following, the results of some numerical experiments are presented, 
in which all the results are the average values obtained over multiple experiments (in all the experiments, training and test datasets were fixed). 
In the following experiments, $S$ (which is the number of sample points for the sampling approximation) was set as follows unless otherwise noted: 
in the learning stage (i.e., in equation (\ref{eqn:grad_likelihood_MDRBM_MCI})) $S = 5$; 
while, in the inference state (i.e., in equation (\ref{eqn:classProb_MDRBM_MCI})) $S = 50$.

\subsection{Experiment for MNIST}
\label{sec:experiment_MNIST}

In the experiments of this section, a dataset obtained from MNIST was used.
MNIST is a database of ten handwritten digits, $0, 1,\ldots,9$, consisting of 70000 data points. 
Each data point in MNIST includes the input data, a $28 \times 28$ digit image, and the corresponding target digit label. 
The sizes of the training and test datasets used in this experiment were $N_{\mrm{train}} = 3000$ and $N_{\mrm{test}} = 10000$, respectively, 
which were randomly selected from the MNIST database.
The size of the training datasets was limited to emphasize the over-fitting problem.

\begin{figure*}[tb]
\centering
\includegraphics[height=4.5cm]{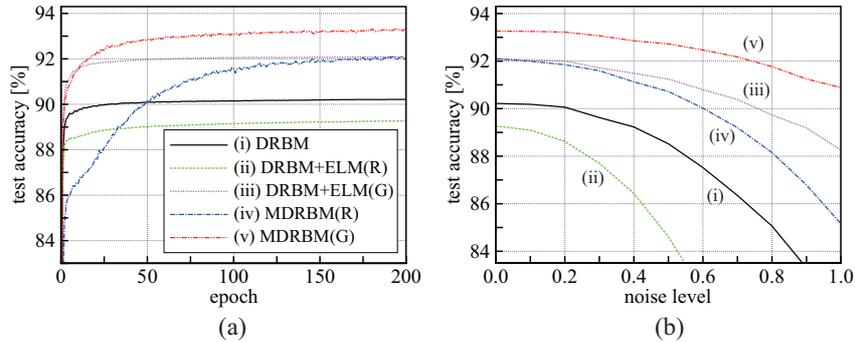}
\caption{(a) Classification accuracy for the (clean) test dataset against the number of training epochs 
and (b) that for the noisy test dataset against the noise level $\sigma$ of AWGN, for MNIST. 
In these experiments, $S = 5$ was used in the inference stage in MDRBM(R) and MDRBM(G).}
\label{fig:N3000_MNIST}
\end{figure*}

Using the size-limited training dataset, five different models, 
(i) DRBM~\cite{DRBM2012} (baseline), (ii) DRBM+ELM(R), (iii) DRBM+ELM(G), (iv) MDRBM(R), and (v) MDRBM(G), were trained. 
(ii) and (iii) are DRBMs with (non-probabilistic) ELM layers in which the values of $\bm{z}$ are deterministically determined by $z_j = \tanh(u_j(\bm{x}))$.
Here, ``(R)'' and ``(G)'' indicate the type of setting of the untrained parameters $\theta_0$. 
In (R), $\bm{b}^{(0)}$ was set to zero, and $\bm{w}^{(0)}$ was randomly drawn from a Gaussian with mean zero and standard deviation $1/\sqrt{n}$. 
On the other hand, in (G), $\theta_0$ was determined from the GBRBM-learning method as discussed in section \ref{sec:Using_GBRBM}. 
(iv) and (v) are the proposed models of this study, where (R) and (G) have the same meanings as in (ii) and (iii), respectively. 
It is noteworthy that (iii) can be viewed as an extension of RBM-ELM~\cite{ELM-GBRBM2018}.
During training, the trained parameters $\theta$ of these five models were initialized using the Xavier initialization~\cite{Xavier2010},  
and the Adam optimizer~\cite{Adam2015} with the mini-batch size of 100 was used. 
The sizes of the layers were $n=784$, $H=\mcal{H} = 500$, and $K = 10$.
Figure \ref{fig:N3000_MNIST}(a) shows the classification accuracies for the test dataset against the number of training epochs. 
It is observed that (v) shows the best accuracy. 

Next, results of a test of noise robustness are presented. 
In the test, additive white Gaussian noise (AWGN) were added to the input data in the test dataset, 
following which the classification accuracies was checked for the ``noisy'' test dataset of the models trained using the ``clean'' training dataset. 
Figure \ref{fig:N3000_MNIST}(b) shows the classification accuracies for the noisy test dataset against the noise level of AWGN.
Here, the noise level represents the standard deviation $\sigma$ of AWGN. 
In this test, the best models in terms of the test accuracy obtained during training were used.
From figure \ref{fig:N3000_MNIST}(b), it is observed that MDRBM(G) is the best and DRBM+ELM(G) is the second best. 
This suggests that the GBRBM-learning method is especially effective for noise robustness, 
which implicitly supports the claim in reference~\cite{ELM-GBRBM2018}.

\subsection{Experiment for benchmark datasets}
\label{sec:experiment_benchmark}

In this section, the promising models, MDRBM(G) and DRBM+ELM(G), are compared with the other existing models, 
CDRBM~\cite{CDRBM2019}, RBM-ELM~\cite{ELM-GBRBM2018}, and a standard four-layered feed-forward neural network (4NN), 
using the experiments of the test of noise robustness. 
The scheme of this test is basically the same as the second experiment in the previous section. 
The noise-robustness test was performed using 
the four different datasets obtained from MNIST, F-MNIST, ULC~\cite{ULC2013}, and CIFAR-10.
The sizes of training and test datasets used in the test were $N_{\mrm{train}} = 3000, 6000, 472, 3000$ and $N_{\mrm{test}} = 10000, 10000, 203, 10000$ 
for MNIST, F-MNIST, ULC, and CIFAR-10, respectively. The sizes of the training datasets were again limited. 
The input RGB color images in CIFAR-10 were converted to the grayscaled images by BT.601.

The settings of the models for the four different datasets were as follows.
For MINST, F-MNIST, and CIFAR-10, the sizes of the two hidden layers were 500. 
For ULC, the sizes of the hidden layers were 100. 
In 4NN, the ReLU activation and the He initialization~\cite{He2015} were used. 
During training of the models, the Adam optimizer was used; the mini-batch size was 100 for MNIST, F-MNIST, and CIFAR-10 
and was 20 for ULC. 

Tables \ref{tab:test_MNIST}--\ref{tab:test_CIFAR10} show the results of the noise-robustness test for the four different datasets. 
The values in the tables are the classification accuracies for the ``noisy'' test datasets with AWGN. 
The columns of $\sigma = 0$ correspond to the accuracies for the ``clean'' test datasets. 
In almost all cases, MDRBM(G) shows the best or the second best accuracies. 
Table \ref{tab:ADR} shows the accuracy-degradation rate (ADR) for each dataset, 
in which ADR [\%] is defined by 
\begin{align*}
\mrm{ADR}:= \frac{[\text{test accuracy}]_{\sigma = 0}-[\text{test accuracy}]_{\sigma = 1}}{[\text{test accuracy}]_{\sigma = 0}}\times 100.
\end{align*}
A model having lower ADR is stronger for noise in the input data.
MDRBM(G) is superior to the other models in terms of ADR. 

For MNIST, the results of MDRBM(G) shown in table \ref{tab:test_MNIST} are different from those shown in figure \ref{fig:N3000_MNIST}(b) 
(the results shown in table \ref{tab:test_MNIST} are better), even through they used the same trained GBRBM.
This accuracy difference comes from the difference in $S$ in the inference stage: $S = 5$ and $S = 50$ were used in figure \ref{fig:N3000_MNIST}(b) 
and in table \ref{tab:test_MNIST}, respectively. 
Empirically, a larger $S$ tends to improve the performance, especially, in the inference stage.

\subsection{Discussion: Effective of PELM layer in inference}
\label{sec:discussion}

In tables \ref{tab:test_MNIST}--\ref{tab:ADR}, MDRBM(G) is superior to DRBM+ELM(G) in almost all cases. 
This suggests that the probabilistic treatment of $\bm{z}$ is important for the performance of the noise robustness. 
In the training stage, this probabilistic treatment functions as a data-augmentation.
As mentioned in section \ref{sec:MDRBM}, this probabilistic treatment is important for not only the training stage but the inference stage. 

This can be understood as follows. 
Consider an input $\mbf{x}$. The stacked DRBM is received $z_j = \tanh(u_j(\mbf{x}))$ in DRBM+EML(G), 
while it is received $z_j$ drawn from the Bernoulli distribution in equation (\ref{eqn:PELM_layer}) in MDRBM(G). 
Suppose that AWGN with noise level $\sigma$, is added to input, $\mbf{x} + \bm{\eta}$, 
and that $u_j(\mbf{x}) = 0$ for the simplicity. 
For the AWGN, the expectations of $z_j$ in DRBM+EML(G) and MDRBM(G) are zero: 
\begin{align*}
\mrm{E}[z_j \mid \sigma^2]=\int_{-\infty}^{\infty} \tanh\Big(\sum_{i=1}^n w_{j,i}^{(0)} \eta_i\Big)\mcal{N}_n(\bm{\eta} \mid \sigma^2\bm{I})d\bm{\eta} = 0,
\end{align*}
where $\mcal{N}_n(\bm{\eta} \mid \sigma^2\bm{I})$ is the $n$-dimensional Gaussian with zero mean vector and covariant (diagonal) matrix $\sigma^2\bm{I}$. 
The variances of $z_j$ in DRBM+EML(G) and MDRBM(G) are 
\begin{align*}
\mrm{V}_{\mrm{ELM}}[z_j \mid \sigma^2]&= \int_{-\infty}^{\infty} \tanh^2\Big(\sum_{i=1}^n w_{j,i}^{(0)} \eta_i\Big)\mcal{N}_n(\bm{\eta} \mid \sigma^2\bm{I})d\bm{\eta},\\
\mrm{V}_{\mrm{PELM}}[z_j \mid \sigma^2]&= 1,
\end{align*}
respectively. It can be proven that $\mrm{V}_{\mrm{ELM}}[z_j \mid \sigma^2]$ is a monotonically increasing function with respect to $\sigma^2$.
Therefore, the variance of $z_j$ in DRBM+EML(G) increases with the increase in the noise level, 
while that in MDRBM(G) is always one for any $\sigma^2$.
This implies that the distribution of $z_j$ of MDRBM(G) is more robust for the noise than that of DRBM+EML(G) in terms of the variance. 

\begin{table}[t]
\caption{Results of the noise-robustness test for MNIST}
\label{tab:test_MNIST}
\centering
\begin{tabular}{lcccccc}
            & \multicolumn{6}{c}{noise level $\sigma$} \\ \cline{2-7} 
            &0    &0.2   &0.4  &0.6  &0.8  &1 \\ \hline
MDRBM(G)    &\textbf{94.5} &\textbf{94.4} &\textbf{94.3} &\textbf{94.0} &\textbf{93.5} &\textbf{92.9} \\ \hline
DRBM+ELM(G) &92.1 &92.0 &91.5 &90.8 &89.7 &88.3 \\ \hline
DRBM        &90.2 &90.1 &89.2 &87.5 &85.1 &81.6 \\ \hline
CDRBM       &89.4 &89.1 &88.2 &86.8 &84.7 &82.1 \\ \hline
RBM-ELM     &87.6 &87.3 &86.6 &85.6 &83.9 &81.7 \\ \hline
4NN         &92.4 &92.2 &91.5 &90.2 &88.1 &85.0 \\ \hline
\end{tabular}
\end{table}

\begin{table}[t]
\caption{Results of the noise-robustness test for F-MNIST}
\label{tab:test_F-MNIST}
\centering
\begin{tabular}{lcccccc}
            & \multicolumn{6}{c}{noise level $\sigma$} \\ \cline{2-7} 
            &0    &0.2   &0.4  &0.6  &0.8  &1 \\ \hline
MDRBM(G)    &85.9 &85.8 &\textbf{85.5} &\textbf{85.0} &\textbf{84.3} &\textbf{83.3} \\ \hline
DRBM+ELM(G) &84.1 &83.9 &83.5 &82.8 &82.0 &80.7 \\ \hline
DRBM        &85.6 &85.3 &84.0 &81.9 &79.0 &75.9 \\ \hline
CDRBM       &84.9 &84.2 &82.6 &80.0 &76.7 &73.1 \\ \hline
RBM-ELM     &81.5 &81.2 &80.3 &79.0 &77.1 &74.7 \\ \hline
4NN         &\textbf{86.6} &\textbf{86.2} &85.2 &83.9 &81.2 &78.3 \\ \hline
\end{tabular}
\end{table}

\begin{table}[t]
\caption{Results of the noise-robustness test for ULC}
\label{tab:test_ULC}
\centering
\begin{tabular}{lcccccc}
            & \multicolumn{6}{c}{noise level $\sigma$} \\ \cline{2-7} 
            &0    &0.2   &0.4  &0.6  &0.8  &1 \\ \hline
MDRBM(G)    &\textbf{78.7} &\textbf{78.4} &\textbf{78.0} &\textbf{77.5} &\textbf{76.7} &\textbf{75.4} \\ \hline
DRBM+ELM(G) &78.6 &77.2 &76.4 &75.7 &74.2 &73.0 \\ \hline
DRBM        &78.5 &76.9 &74.9 &72.6 &69.3 &66.2 \\ \hline
CDRBM       &78.3 &77.0 &75.7 &74.6 &72.5 &70.2 \\ \hline
RBM-ELM     &73.9 &71.8 &67.7 &62.4 &56.8 &49.6 \\ \hline
4NN         &77.7 &76.0 &74.1 &71.2 &66.8 &62.3 \\ \hline
\end{tabular}
\end{table}

\begin{table}[t]
\caption{Results of the noise-robustness test for CIFAR-10}
\label{tab:test_CIFAR10}
\centering
\begin{tabular}{lcccccc}
            & \multicolumn{6}{c}{noise level $\sigma$} \\ \cline{2-7} 
            &0    &0.2   &0.4  &0.6  &0.8  &1 \\ \hline
MDRBM(G)    &30.3 &30.2 &30.0 &30.0 &\textbf{29.7} &\textbf{29.3} \\ \hline
DRBM+ELM(G) &28.1 &27.9 &27.8 &27.6 &27.1 &26.6 \\ \hline
DRBM        &27.3 &27.0 &26.4 &25.1 &23.9 &22.7 \\ \hline
CDRBM       &27.4 &27.0 &26.4 &25.3 &24.4 &23.3 \\ \hline
RBM-ELM     &22.5 &22.2 &21.9 &21.1 &20.4 &19.6 \\ \hline
4NN         &\textbf{33.3} &\textbf{32.8} &\textbf{31.9} &\textbf{30.3} &28.7 &27.1 \\ \hline
\end{tabular}
\end{table}

\begin{table}[t]
\caption{Accuracy degradation rates for the benchmark datasets}
\label{tab:ADR}
\centering
\begin{tabular}{lcccc}
            & \multicolumn{4}{c}{dataset}      \\ \cline{2-5} 
            & MNIST & F-MNIST & UCL   & CIFAR-10 \\ \hline
MDRBM(G)    & \textbf{1.8}   & \textbf{3.1}     & \textbf{4.2}   & \textbf{3.1}      \\ \hline
DRBM+ELM(G) & 4.1   & 4.0     & 7.1   & 5.3      \\ \hline
DRBM        & 9.5   & 11.3    & 15.7  & 16.8     \\ \hline
CDRBM       & 8.2   & 13.9    & 10.3  & 15.0     \\ \hline
RBM-ELM     & 6.7   & 8.3     & 32.9  & 12.9     \\ \hline
4NN         & 8.0   & 9.6     & 19.8  & 18.6     \\ \hline
\end{tabular}
\end{table}

\section{Conclusion and Future Works}
\label{sec:conclusion}

In this paper, a probabilistic four-layered neural network named MDRBM was proposed 
by combining DRBM and a PELM layer. 
The PELM layer is an untrained probabilistic layer inspired by ELM. 
Unlike DBM, the learning of the proposed MDRBM can be performed by applying only a simple sample approximation. 
Moreover, we proposed an effective setting of the parameters in the PELM layer based on the GBRBM-learning method, 
which is in accordance with a similar idea proposed in reference~\cite{ELM-GBRBM2018}. 
In the numerical experiments using some benchmark datasets, MDRBM with the GBRBM-learning method (i.e., MDRBM(G)) 
largely improves the property of the noise robustness against the input noise.

In the proposed model, the quality of the GBRBM-learning is essentially important.
Karakida \textit{et. al.} proposed an efficient learning algorithm for GBRBM~\cite{Karakida2016} 
that enables the exact training of GBRBM by limiting the weight parameters $\bm{w}^{(0)}$ in the Stiefel manifold. 
In GBRBM trained using this method, the hidden variables $\bm{z}$ are statistically independent of each other; 
therefore, they will produce a feature mapping similar to that of principal component analysis. 
Applying this learning method to MDRBM is one of the most interesting future works.



\section*{Acknowledgment}
This work was partially supported by JSPS KAKENHI (Grant Numbers 18K11459 and 18H03303), 
JST CREST (Grant Number JPMJCR1402), and the COI Program of JST (Grant Number JPMJCE1312).

\bibliographystyle{IEEEtran}
%

\bibliography{citation}

\begin{thebibliography}{10}
\providecommand{\url}[1]{#1}
\csname url@samestyle\endcsname
\providecommand{\newblock}{\relax}
\providecommand{\bibinfo}[2]{#2}
\providecommand{\BIBentrySTDinterwordspacing}{\spaceskip=0pt\relax}
\providecommand{\BIBentryALTinterwordstretchfactor}{4}
\providecommand{\BIBentryALTinterwordspacing}{\spaceskip=\fontdimen2\font plus
\BIBentryALTinterwordstretchfactor\fontdimen3\font minus
  \fontdimen4\font\relax}
\providecommand{\BIBforeignlanguage}[2]{{%
\expandafter\ifx\csname l@#1\endcsname\relax
\typeout{** WARNING: IEEEtran.bst: No hyphenation pattern has been}%
\typeout{** loaded for the language `#1'. Using the pattern for}%
\typeout{** the default language instead.}%
\else
\language=\csname l@#1\endcsname
\fi
#2}}
\providecommand{\BIBdecl}{\relax}
\BIBdecl

\bibitem{DRBM2012}
H.~Larochelle, M.~Mandel, R.~Pascanu, and Y.~Bengio, ``Learning algorithms for
  the classification restricted boltzmann machine,'' \emph{The Journal of
  Machine Learning Research}, vol.~13, no.~1, pp. 643--669, 2012.

\bibitem{RBM1986}
P.~Smolensky, ``Information processing in dynamical systems: foundations of
  harmony theory,'' \emph{Parallel distributed processing: Explorations in the
  microstructure of cognition}, vol.~1, pp. 194--281, 1986.

\bibitem{CD2002}
G.~E. Hinton, ``Training products of experts by minimizing contrastive
  divergence,'' \emph{Neural Computation}, vol.~14, no.~8, pp. 1771--1800,
  2002.

\bibitem{DBM2009}
R.~Salakhutdinov and G.~E. Hinton, ``Deep boltzmann machines,'' \emph{In Proc.
  of the 12th International Conference on Artificial Intelligence and
  Statistics}, pp. 448--455, 2009.

\bibitem{DeepGBRBM2013}
K.~Cho, T.~Raiko, and A.~Ilin, ``Gaussian-bernoulli deep boltzmann machine,''
  \emph{In Proc. of the 2013 International Joint Conference on Neural
  Networks}, pp. 1--7, 2013.

\bibitem{DL2016}
I.~Goodfellow, Y.~Bengio, and A.~Courville, \emph{Deep Learning}.\hskip 1em
  plus 0.5em minus 0.4em\relax MIT Press, 2016.

\bibitem{ELM2006}
G.~Huang, Q.~Zhu, and C.~Siew, ``Extreme learning machine: Theory and
  applications,'' \emph{Neurocomputing}, vol.~70, pp. 489--501, 2006.

\bibitem{RBM-Sci2006}
G.~E. Hinton and R.~Salakhutdinov, ``Reducing the dimensionality of data with
  neural networks,'' \emph{Science}, vol. 313, no. 5786, pp. 504--507, 2006.

\bibitem{GBRBM2011}
K.~Cho, A.~Ilin, and T.~Raiko, ``Improved learning of gaussian-bernoulli
  restricted boltzmann machines,'' \emph{In Proceedings of the 12th
  International Conference on Artificial Neural Networks}, pp. 10--17, 2011.

\bibitem{ELM-GBRBM2018}
A.~G. Pacheco, R.~A. Krohling, and C.~A. da~Silva, ``Restricted boltzmann
  machine to determine the input weights for extreme learning machines,''
  \emph{Expert Systems with Applications}, vol.~96, pp. 77--85, 2018.

\bibitem{CDRBM2019}
Y.~Yokoyama, T.~Katsumata, and M.~Yasuda, ``Restricted boltzmann machine with
  multivalued hidden variables: a model suppressing over-fitting,'' \emph{The
  Review of Socionetwork Strategies}, vol.~13, no.~2, pp. 253--266, 2019.

\bibitem{ULC2013}
B.~Johnson and Z.~Xie, ``Classifying a high resolution image of an urban area
  using super-object information,'' \emph{ISPRS Journal of Photogrammetry and
  Remote Sensing}, vol.~83, pp. 40--49, 2013.

\bibitem{Bishop2006}
C.~M. Bishop, \emph{Pattern Recognition and Machine Learning}.\hskip 1em plus
  0.5em minus 0.4em\relax Springer-Verlag New York, 2006.

\bibitem{Xavier2010}
X.~Glorot and Y.~Bengio, ``Understanding the difficulty of training deep
  feedforward neural networks,'' \emph{In Proc. of the 13th International
  Conference on Artificial Intelligence and Statistics}, vol.~9, pp. 249--256,
  2010.

\bibitem{Adam2015}
D.~P. Kingma and L.~J. Ba, ``Adam: A method for stochastic optimization,''
  \emph{In Proc. of the 3rd International Conference on Learning
  Representations}, pp. 1--13, 2015.

\bibitem{He2015}
K.~He, X.~Zhang, S.~Ren, and J.~Sun, ``Delving deep into rectifiers: Surpassing
  human-level performance on imagenet classification,'' \emph{In Proc. of the
  2015 IEEE International Conference on Computer Vision}, pp. 1026--1034, 2015.

\bibitem{Karakida2016}
R.~Karakida, M.~Okada, and S.~Amari, ``Maximum likelihood learning of rbms with
  gaussian visible units on the stiefel manifold,'' \emph{In Proc. of the 24th
  European Symposium on Artificial Neural Networks, Computational Intelligence
  and Machine Learning}, pp. 159--164, 2016.

\end{thebibliography}

\end{document}